\documentclass[runningheads]{llncs}
\usepackage{graphicx}
\usepackage{booktabs}
\usepackage{pifont}
\usepackage{amsmath}
\usepackage{amsfonts}
\usepackage{amssymb}
\usepackage{listings}
\usepackage{subfigure}
\usepackage{multirow}

\usepackage{microtype}

\usepackage{changepage}
\usepackage{lipsum}
\usepackage{tabularx}

\newlength{\offsetpage}
{\end{adjustwidth}}

\newcommand{\NAME}{\textsc{Plumber}}
\newcommand{\qq}[1]{``#1''}

\newcommand{\ie}{i.e.,~}
\newcommand{\eg}{e.g.,~}
\newcommand{\componentsCount}{33~}
\newcommand{\pipelinesCount}{264~}

\newcommand{\cf}{cf.~}
\newcommand{\rNum}[1]{\expandafter{\romannumeral #1\relax}}
\newcommand{\RNum}[1]{\uppercase\rNum{#1}}
\newcommand{\rarrow}{\ensuremath{\rightarrow}}

\begin{document}
\title{Better Call the Plumber: Orchestrating Dynamic Information Extraction Pipelines}
\titlerunning{Better Call the Plumber}
\author{Mohamad Yaser Jaradeh\inst{1}\orcidID{0000-0001-8777-2780} \and
Kuldeep Singh\inst{2}\orcidID{0000-0002-5054-9881} \and
Markus Stocker\inst{3}\orcidID{0000-0001-5492-3212} \and
Andreas Both\inst{4}\orcidID{0000-0002-9177-5463} \and
S\"oren Auer\inst{3}\orcidID{0000-0002-0698-2864}}
\authorrunning{Jaradeh et al.}
\institute{L3S Research Center, Leibniz University Hannover, Germany
\email{jaradeh@l3s.de} \and
Zerotha-Research \& Cerence GmbH, Germany
\email{kuldeep.singh1@cerence.com} \and
TIB Leibniz Information Centre for Science and Technology, Germany
\email{\{markus.stocker, auer\}@tib.eu} \and
Anhalt University of Applied Sciences, Germany
\email{andreas.both@hs-anhalt.de}
}
\maketitle         
\begin{abstract}
We propose \NAME, the first framework that brings together the research community’s disjoint information extraction (IE) efforts. The \NAME\ architecture comprises \componentsCount reusable components for various Knowledge Graphs (KG) information extraction subtasks, such as coreference resolution, entity linking, and relation extraction. Using these components, \NAME\ dynamically generates suitable information extraction pipelines and offers overall \pipelinesCount distinct pipelines. We study the optimization problem of choosing suitable pipelines based on input sentences. To do so, we train a transformer-based classification model that extracts contextual embeddings from the input and finds an appropriate pipeline. We study the efficacy of \NAME\ for extracting the KG triples using standard datasets over two KGs: DBpedia, and Open Research Knowledge Graph (ORKG). Our results demonstrate the effectiveness of \NAME\ in dynamically generating KG information extraction pipelines, outperforming all baselines agnostics of the underlying KG. Furthermore, we provide an analysis of collective failure cases, study the similarities and synergies among integrated components, and discuss their limitations.

\end{abstract}
\section{Introduction and Motivation}\label{sec:intro}
\begin{figure}[b]
    \centering
    \includegraphics[width=.8\columnwidth]{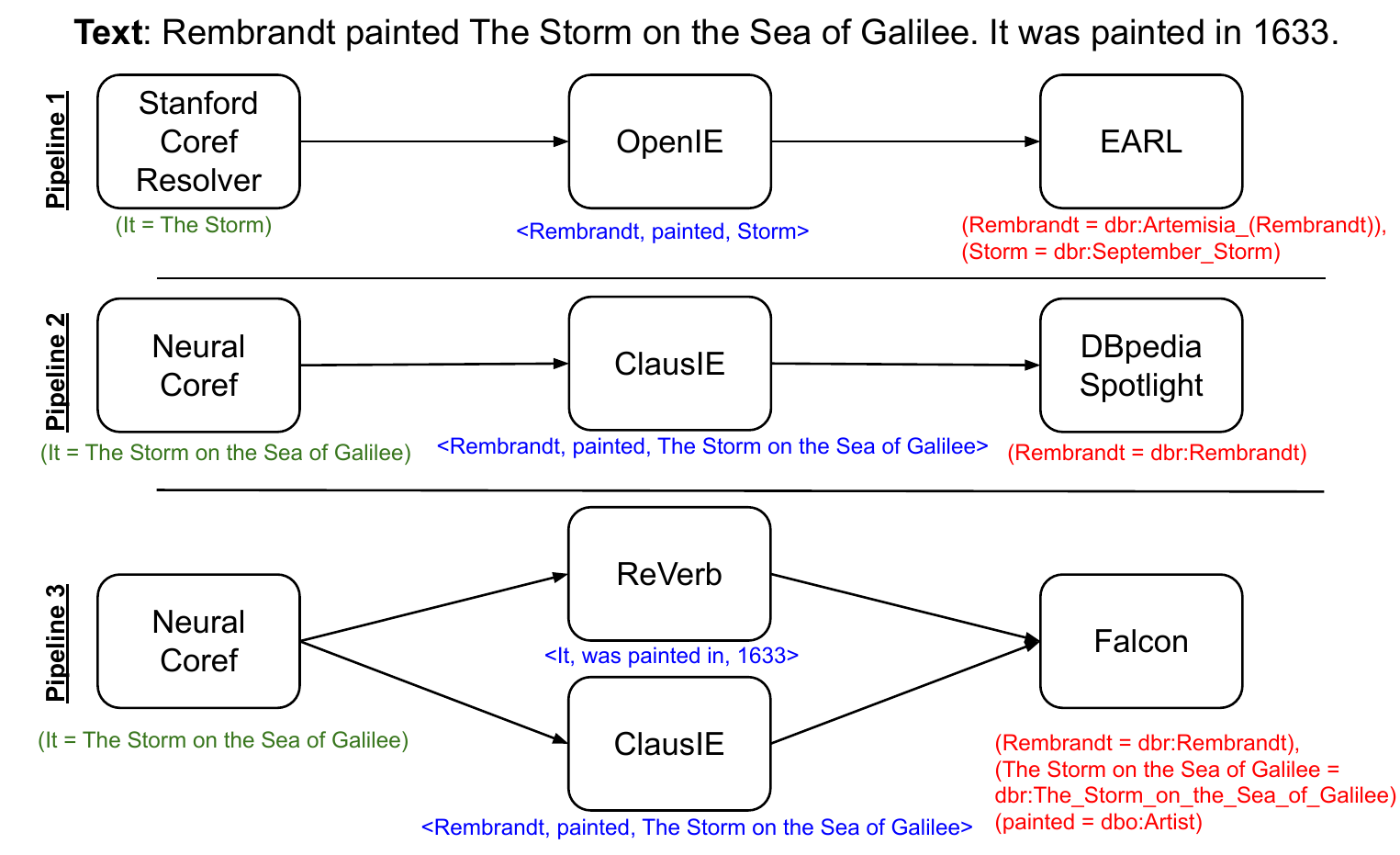}
    \caption{Three example information extraction pipelines showing different results for the same text snippet. Each pipeline consists of coreference resolution, triple extractors, and entity/relation linking components.}
    \label{fig:example}
\end{figure}

In last one decade, publicly available KGs (DBpedia~\cite{dbpedia} and Wikidata~\cite{wikidata} ) have become rich sources of structured content used in various applications, including Question Answering (QA), fact checking, and dialog systems~\cite{frankenstein,bastos2020recon}. 
The research community developed numerous approaches to extract triple statements~\cite{kgBert}, keywords/topics~\cite{hierarchical-topics}, tables~\cite{tablepedia,briQ,TDMS-IE}, or entities~\cite{falcon,falcon2} from unstructured text to complement KGs. Despite extensive research, public KGs are not exhaustive and require continuous effort to align newly emerging unstructured information to the concepts of the KGs. \\
\textbf{Research Problem:} This work was motivated by an observation of recent approaches~\cite{falcon,tablepedia,earl} that automatically align unstructured text to structured data on the Web. Such approaches are not viable in practice for extracting and structuring information because they only address very specific subtasks of the overall KG information extraction problem. If we consider the exemplary sentence \textit{Rembrandt painted The Storm on the Sea of Galilee. It was painted in 1633.} (\cf Figure~\ref{fig:example}). To extract statements aligned with the DBpedia KG from the given sentences, a system must first recognize the entities and relation surface forms in the first sentence. The second sentence requires an additional step of the coreference resolution, where \emph{It} must be mapped to the correct entity surface form (namely, \emph{The Storm on the Sea of Galilee}). The last step requires the mapping of entity and relation surface forms to the respective DBpedia entities and predicates. 
There has been extensive research in aligning concepts in unstructured text to KG, including entity linking~\cite{earl,tagme}, relation linking~\cite{falcon2,relmatch,bastos2020recon}, and triple classification \cite{dong2019triple}.
However, these efforts are disjoint, and little has been done to align unstructured text to the complete KG triples (\ie represented as subject, predicate, object)~\cite{t2kg}. Furthermore, many entity and relation linking tools have been reused in pipelines of QA systems~\cite{frankenstein,kimokbqa}. The literature suggests that once different approaches put forward by the research community are combined, the resulting pipeline-oriented integrated systems can outperform monolithic end-to-end systems \cite{liang2020querying}. 
For the KG information extraction task, however, to the best of our knowledge, approaches aiming at dynamically integrating and orchestrating various existing components do not exist. \\
\textbf{Objective and Contributions:} Based on these observations, we build a framework that enables the integration of previously disjoint efforts on the KG-IE task under a single umbrella. 
We present the \NAME\ framework (\cf Figure~\ref{fig:architecture}) for creating Information Extraction pipelines. \NAME\ integrates \componentsCount reusable components released by the research community for the subtasks entity linking (EL), relation linking (RL), text triple extraction (TE) (subject, predicate, object), and coreference resolution (CR). Overall, there are \pipelinesCount different composable KG information extraction pipelines (generated by the possible combination of the available \componentsCount components, \ie for DBpedia 3 CRs, 8 TEs, 10 EL/RLs gives 3*8*10=240, and 4*3*2=24 for the ORKG. Hence, 240+24=\pipelinesCount pipelines). \NAME\ implements a transformer-based classification algorithm that intelligently chooses the best pipeline based on the unstructured input text.

We perform an exhaustive evaluation of \NAME\ on the two large-scale KGs DBpedia, and Open Research Knowledge Graph (ORKG)~\cite{orkg} to investigate the efficacy of \NAME\ in creating KG triples from unstructured text. We demonstrate that independent of the underlying KG; \NAME\ can find and assemble different extraction components to produce better suited KG triple extraction pipelines, significantly outperforming existing baselines.
In summary, we provide the following novel contributions: \rNum{1}) The \NAME\ framework is the first of its kind for dynamically assembling and evaluating information extraction pipelines based on sequence classification techniques and for a given input text. \NAME\ is easily extensible and configurable, thus enabling the rapid creation and adjustment of new information extraction components and pipelines. Researchers can also use the framework for running IE components independently for specific subtasks such as triple extraction and entity linking. \rNum{2}) A collection of \componentsCount reusable IE components that can be combined to create \pipelinesCount distinct IE pipelines. \rNum{3}) The exhaustive evaluation and our detailed ablation study of the integrated components and composed pipelines on various input text will guide future research for collaborative KG information extraction. 

We motivate our work with a running example; the sentence \emph{Rembrandt painted The Storm on the Sea of Galilee. It was painted in 1633}.
Multiple steps are required to extract these formally represented statements from the given text.
First, the pronoun \emph{it} in the second sentence should be replaced by \emph{The Storm on the Sea of Galilee} using a coreference resolver.
Next, a triple extractor should extract the correct text triples from the natural language text,
\ie \texttt{<Rembrandt, painted, The Storm on the Sea of Galilee>}, and \texttt{<The Storm on the Sea of Galilee, painted in, 1633>}.
In the next step, the entity and relation linking component aligns the entity and relation surface forms extracted in the previous step to the DBpedia entities: \texttt{dbr:Rembrandt} for \emph{Rembrandt van Rijn}, and \texttt{dbr:The\_Storm\_on\_the\_Sea\_of\_Galilee} for \emph{The Storm on the Sea of Galilee}, and for relations: \texttt{dbo:Artist} for \emph{painted}, and \texttt{dbp:year} for \emph{painted in}. Figure~\ref{fig:example} illustrates our running example and shows three \NAME\ IE pipelines with different results.
In Pipeline~1, the coreference resolver is unable to map the pronoun \emph{it} to the respective entity in the previous sentence.
Moreover, the triple extractor generates incomplete triples, which also hinders the task of the entity and relation linker in the last step.
Pipeline~2 uses a different set of components, and its output differs from the first pipeline.
Here, the coreference resolution component is able to correctly co-relate the pronoun \emph{it} to \emph{The Storm on the Sea of Galilee}, and extract the text triple correctly. However, the overall result is only partially correct because the second triple is not extracted.
Also, the linking component is not able to spot the second entity. 
Pipeline~3 correctly extracts both triples. This pipeline employs the same component as the second pipeline for coreference resolution but also includes an additional information extraction component (\ie ReVerb~\cite{reverb}) and a joint entity and relation linking component, namely Falcon~\cite{falcon}. With this combination of components, the text triple extractors were able to compensate for the loss of information in the second pipeline by adding one more component. Using the extracted text triples, the last component of the pipeline, a joint entity and relation linking tool, can map both triple components correctly to the corresponding KG entities. 

The reminder of this article is organized as follows. Related work is reviewed in Section~\ref{sec:related}. Section~\ref{sec:plumber} presents \NAME, which is extensively evaluated in Section~\ref{sec:eval}. Section~\ref{sec:discussion} discusses the results, and Section~\ref{sec:conclusion} concludes and outlines directions for future research and work.

\section{Related Work}
\label{sec:related}
In the last decade, many open source tools have been released by the research community to tackle IE tasks for KGs. These IE components are not only used for end-to-end KG triple extraction but also for various other tasks, such as:\\
\textbf{Text Triple Extraction}: The task of open information extraction is a well studied researched task in the NLP community~\cite{openie}.
It relies on NER (Named Entity Recognition) and RE (Relation Extraction).
SalIE~\cite{salIE} uses MinIE~\cite{minie} in combination with PageRank and clustering to find facts in the input text.
Furthermore, OpenIE~\cite{openie} leverages linguistic structures to extract self-contained clauses from the text. A comprehensive survey by Niklaus et al.~\cite{niklaus2018survey} provides detailed about such techniques.\\
\textbf{Entity and Relation Linking}:
Entity and relation linking is a widely studied researched topic in the NLP, Web, and Information Retrieval research communities~\cite{Balog2018,bastos2020recon,opentapioca}. Often, entity and relation linking is performed independently. DBpedia Spotlight~\cite{spotlight} is one of the first approaches for entity recognition and disambiguation over DBpedia. TagMe~\cite{tagme} links entities to DBpedia using in-link matching to disambiguate candidates entities. Others tools such as RelMatch~\cite{relmatch} do not perform entity linking and only focus on linking the relation in the text to the corresponding KG relation. Recon~\cite{bastos2020recon} uses graph neural networks to to map relations between the entities with the assumption that entities are already linked in the text. EARL~\cite{earl} is a joint linking tool over DBpedia and models the task as a generalized traveling salesperson problem. Sakor et al.~\cite{falcon} proposed Falcon, a linguistic rules based tool for joint entity and relation linking over DBpedia. \\
\textbf{Coreference Resolution}: This task is used in conjunction with other tasks in NLP pipelines to disambiguate text and resolve syntactic complexities. The Stanford Coreference Resolver~\cite{stanford_coref} uses a multi pass sieve of deterministic coreference models. Clark and Manning~\cite{neuralcoref} use reinforcement learning to fine-tune a neural mention-ranking model for coreference resolution. And more recently~\cite{hmtl}.\\
\textbf{Frameworks and Dynamic Pipelines}:
There have been few attempts in various domains aiming to consolidate the disjoint efforts of the research community under a single umbrella for solving a particular task. The Gerbil platform~\cite{usbeck2015gerbil} provides an easy-to-use web-based platform for the agile comparison of entity linking tools using multiple datasets and uniform measuring approaches. OKBQA~\cite{kimokbqa} is a community effort for the development of multilingual open knowledge base and QA systems. Frankenstein integrates 24 QA components to build QA systems collaboratively on-top of the Qanary integration framework~\cite{DBLP:conf/esws/BothDSSC016}. Other ETL pipelines system exists such as Apache NiFi. Semantic Web Pipes~\cite{semantic-web-pipes} and LarKC~\cite{larkc} are other prominent examples.
\\
\textbf{End-to-End Extraction Systems}:
More recently, end-to-end systems are gaining more attention due to the boom of deep learning techniques. Such systems draw on the strengths of deep models and transformers~\cite{bert,roberta}.
Kertkeidkachorn and Ichise~\cite{t2kg} present an end-to-end system to extract triples and link them to DBpedia. Other attempts such as KG-Bert~\cite{kgBert} leverage deep transformers (\ie BERT~\cite{bert}) for the triple classification task, given the entity and relation descriptions of a triple. KG-Bert does not attempt end-to-end alignment of KG triples from a given input text. Liu et al.~\cite{Seq2RDF} design an encoder-decoder framework with an attention mechanism to extract and align triples to a KG.

\section{Dynamic Information Extraction Pipelining Framework}
\label{sec:plumber}
\begin{figure*}[t]
    \vspace*{-\baselineskip}
	\centering
	\includegraphics[width=\textwidth]{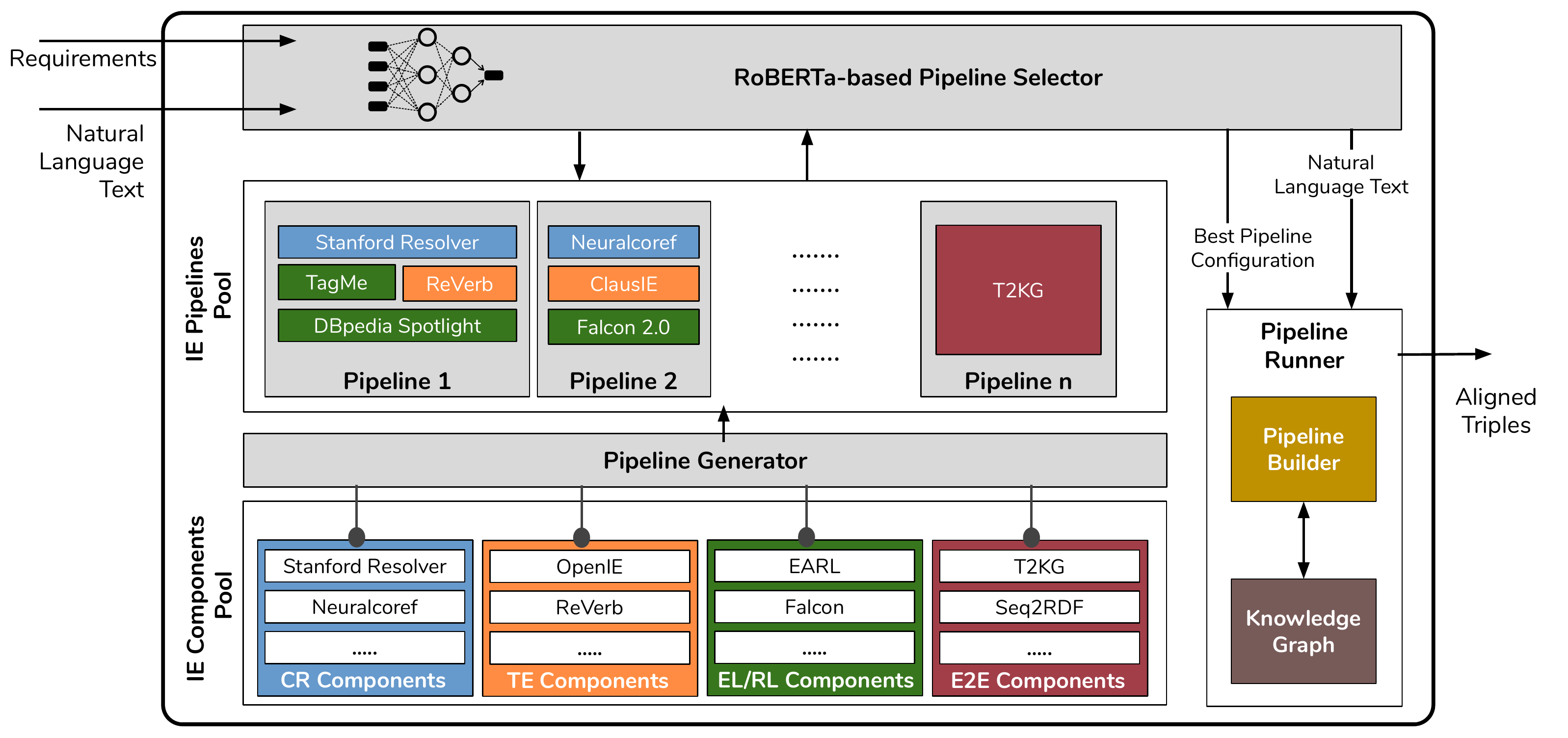}
	\caption{Overview of \NAME's architecture highlighting the components for pipeline generation, selection, and execution. \NAME\ receives an input sentence and requirement (underlying KG) from the user. The framework intelligently selects a suitable pipeline based on the contextual features captured from the input sentence.}
	\label{fig:architecture}
	\vspace*{-\baselineskip}
\end{figure*}

\NAME\ has a modular design (see Figure~\ref{fig:architecture}) where each component is integrated as a microservice. To ensure a consistent data exchange between components, the framework maps the output of each component to a homogeneous data representation using the Qanary~\cite{DBLP:conf/esws/BothDSSC016} methodology. \NAME\ follows three design principles of \rNum{1}) \textit{Isolation}, \rNum{2}) \textit{Reusability}, and \rNum{3}) \textit{Extensibility} inspired by~\cite{frankenstein,usbeck2015gerbil}.

{\bf Dynamic pipeline selection}:
\NAME\ uses a RoBERTa~\cite{roberta} based classifier that given a text and a set of requirements, \NAME\ predicts the best pipeline to extract KG triples. The RoBERTa model acts as intermediary that classifies the contextual embeddings extracted from the input text into a class which represents one of the possible pipelines. Regarding RoBERTa's training, we run each input sequence on all possible pipelines and compute the evaluation metrics F1-score (\ie estimated performance). RoBERTa is fed with the sentence and the sentence-level performance with the best value among all pipelines as the target class. Hence, in practice, the user points \NAME~to a piece of text and internally it uses RoBERTa to classify the text to a class (\ie the pipeline) to execute against the input text.

{\bf Architecture}:
\NAME\ includes the following modules:
\textbf{\rNum{1}) IE Components Pool}: All information extraction components that are integrated within the framework are parts of the pool. 
The components are divided based on their respective tasks, \ie coreference resolution, text triple extraction, as well as entity and relation linking.
\textbf{\rNum{2}) Pipeline Generator}: This module creates possible pipelines depending on the requirements of the components (\ie the underlying KG). Users can manually select the underlying KG and, using the metadata associated with each component, \NAME\ aggregates the components for the concerned KG.
\textbf{\rNum{3}) IE Pipelines Pool}: \NAME\ stores the configurations of the possible pipelines in the pool of pipelines for faster retrieval and easier interaction with other modules.
\textbf{\rNum{4}) Pipeline Selector}: Based on the requirements (\ie underlying KG) and the input text, a RoBERTa based model extracts contextual embeddings from the text and classifies the input into one of the possible classes. Each class corresponds to one pipeline configuration that is held in the IE pipelines pool.
\textbf{\rNum{5}) Pipeline Runner}: Given the input text, and the generated pipeline configuration, the module executes the pipeline and produce the final KG triples.

\section{Evaluation}
\label{sec:eval}
In this section, we detail the empirical evaluation of the framework in comparison to baselines on different datasets and knowledge graphs. As such, we study the following research question: \textit{How does the dynamic selection of pipelines based on the input text affect the end-to-end information extraction task?} 

\subsection{Experimental Setup}
\label{ssec:expsetup}
\paragraph{\textbf{Knowledge Graphs}}
To study the effectiveness of \NAME\ in building dynamic KG information extraction pipelines, we use the following KGs during our evaluation:\\
{\bf DBpedia}~\cite{dbpedia} is containing information extracted automatically from Wikipedia info boxes. DBpedia consists of approximately 11.5B triples~\cite{falcon}.\\
{\bf Open Research Knowledge Graph}~\cite{orkg} (ORKG) collects structured scholarly knowledge published in research articles, using crowd sourcing and automated techniques. In total, ORKG consists of approximately 984K triples.
\paragraph{\textbf{Datasets}}
Throughout our evaluation, we employed a set of existing and newly created datasets for structured triple extraction and alignment to knowledge graphs: the WebNLG~\cite{webnlg} dataset for DBpedia, and COV-triples for ORKG.
~\\
{\bf WebNLG} is the Web Natural Language Generation Challenge. The challenge introduced the task of aligning unstructured text to DBpedia.
In total, the dataset contains 46K triples with 9K triples in the testing and 37K in the training set.
~\\
{\bf COV-triples} is a handcrafted dataset that focuses on COVID-19 related scholarly articles.
The COV-triples dataset consists of 21 abstracts from peer-reviewed articles and aligns the natural language text to the corresponding KG triples into the ORKG.
Three Semantic Web researchers verified annotation quality, and triples approved by all three researchers are part of the dataset.
The dataset contains only 75 triples. Hence, we use the WebNLG dataset for training, and 75 triples are used as a test set. \\
\textbf{Components and Implementation}
The \NAME\ framework integrates \componentsCount components, the components span different IE tasks from Triple Extraction, Entity and Relation Linking, and Coreference Resolution. Most of the components used are open-sourced and they have been evaluated and used by the community in their respective publications. \NAME's code and all related resources are publicly available online at \url{https://git.io/JtT1s}.\\
\textbf{Baselines}
We include the following baselines:~\\
{\bf T2KG}~\cite{t2kg} is an end-to-end static system aligns a given natural language text to DBpedia KG triples.
~\\
{\bf Frankenstein}~\cite{frankenstein} dynamically composes Question Answering pipelines over the DBpedia KG. It employs logistic regression based classifiers for each component for predicting the accuracy and greedily composes a dynamic pipeline of the best components per task. We adapted Frankenstein for the KG information extraction over DBpedia.

\subsection{Experiments}
\label{ssec:experiments}
The section summarizes a variety of experiments to compare the \NAME\ framework against other baselines. 
Note, that evaluating the performance of individual components or their combination is out of this evaluation's scope, since they were already used, benchmarked, and evaluated in the respective publications. We report values of the standard metrics Precision (P), Recall (R), and F1 score (F1). In all experiments, end-to-end components (\eg T2KG) are not part of \NAME. 
\paragraph{\textbf{Performance of Static Pipelines}} In this experiment, we report results of the static pipelines, \ie no dynamic selection of a pipeline based on the input text is considered. We ran all \pipelinesCount pipelines and Table~\ref{tab:pipeline-dynamic-performance} (T2KG \& Static noted rows) reports the performance of the best \NAME\ pipeline against the baselines. \NAME\ static pipeline for DBpedia comprises of NeuralCoref~\cite{neuralcoref} for coreference resolution, OpenIE~\cite{openie} for text triple extraction, TagMe~\cite{tagme} for EL, and Falcon~\cite{falcon} for RL tasks.
Also, in case of Frankenstein, we choose its best performing static pipeline. Results illustrated in the Table~\ref{tab:pipeline-dynamic-performance} confirm that the static pipeline composed by the components integrated in \NAME\ outperforms all baselines on DBpedia. We observe that the performance of pipeline approaches is better than an end-to-end monolithic information extraction approaches. Although the \NAME\ pipeline outperforms the baselines, the overall performance is relatively low. All our components have been trained on distinct corpora in their respective publications and our aim was to put them together to understand their collective strengths and weaknesses. Note, Frankenstein addresses the QA pipeline problem and not all components are comparable and can be applied in the context of information extraction. Thus, we integrated NeuralCoref coreference resolution component and OpenIE triple extraction component used in \NAME\ static pipeline into Frankenstein for providing the same experimental settings.

\paragraph{\textbf{Static Pipeline for Scholarly KG}}
In order to assess how \NAME\ performs on domain-specific use cases, we evaluate the static pipelines' performance on a scholarly knowledge graph. We use the COV-triples dataset for ORKG. To the best of our knowledge, no baseline exists on information extractions of research contribution descriptions over ORKG. Hence, we execute all static pipelines in \NAME\ tailored to ORKG to select the best one as shown in Table~\ref{tab:pipeline-dynamic-performance} (COV-triples rows). \NAME\ pipelines over ORKG extract statements determining the reproductive number estimates for the COVID-19 infectious disease from scientific articles as shown below. 

\begin{lstlisting}[basicstyle=\scriptsize\ttfamily\centering]
@prefix orkg: <http://orkg.org/orkg/resource/>.
@prefix orkgp: <http://orkg.org/orkg/property/>.

orkg:R48100   orkgp:P16022   "2.68" .
\end{lstlisting}
In this example, \emph{orkg:R48100} refers to the  city of Wuhan in China in the ORKG and \emph{orkgp:P16022} is the property ``has R0 estimate (average)''. The number ``2.68'' is the reproductive number estimate.

\paragraph{\textbf{Comparison of the Classification Approaches for Dynamic Pipeline Selection}} 
In this experiment, we study the effect of the transformer-based pipeline selection approach implemented in \NAME\ against the pipeline selection approach of Frankenstein. For a comparable experimental setting, we re-use Frankenstein's classification approach in \NAME, keeping the underlying components precisely the same. 
We perform a 10-fold cross-validation for the classification performance of the employed approach.
Table~\ref{tab:pipeline-selection} indicates that the \NAME\ pipeline selection significantly outperforms baselines across the board.

\begin{table}[t]
\caption{10-fold CV of pipeline selection classifiers wrt. Precision, Recall, and F1 score.}
\centering
\resizebox{.75\columnwidth}{!}{
\begin{tabular}{lccccc}
\hline
\multirow{2}{*}{\textbf{\begin{tabular}[c]{@{}c@{}}Pipeline Selection\\ Approach\end{tabular}}} &
  \multirow{2}{*}{\textbf{Dataset}} &
  \multirow{2}{*}{\textbf{\begin{tabular}[c]{@{}c@{}}Knowledge\\ Graph\end{tabular}}} &
  \multicolumn{3}{c}{\textbf{Classification}} \\ \cline{4-6} 
                              &             &      & \textbf{P} & \textbf{R} & \textbf{F1} \\ \hline
\multirow{2}{*}{Frankenstein~\cite{frankenstein}} & WebNLG      & DBpedia  & 0.732      & 0.751      & 0.741       \\
                             & COV-triples & ORKG & 0.832      & 0.858      & 0.845       \\ \hline
\multirow{2}{*}{\NAME}      & WebNLG      & DBpedia  & \textbf{0.877}      & \textbf{0.900}      & \textbf{0.888}       \\
                              & COV-triples & ORKG & \textbf{0.901}      & \textbf{0.917}      & \textbf{0.909}       \\ \hline
\end{tabular}
}
\label{tab:pipeline-selection}
\end{table}

\paragraph{\textbf{Performance Comparison for KG Information Extraction Task}} 
Our third experiment focuses on comparing the performance of \NAME\ against previous baselines for an end-to-end information extraction task. The results in Table~\ref{tab:pipeline-dynamic-performance} illustrate that the dynamic pipelines built using \NAME\ for KG information extraction outperform the best static pipelines of \NAME\ as well as the dynamically selected pipelines by Frankenstein (rows noted with dynamic). The end-to-end baselines, such as Kertkeidka-chorn and Ichise~\cite{t2kg}. We also observe that in cross-domain experiments for COV-triples datasets, dynamically selected pipelines perform better than the static pipeline. In the cross-domain experiment, the static and dynamic \NAME\ pipelines are relatively better performing than the other two KGs. Unlike components for DBpedia, components integrated into \NAME\ for ORKG are customized for KG triple extraction. We conclude that when components are integrated into a framework such as \NAME\ aiming for the KG information extraction task, it is crucial to select the pipeline based on the input text dynamically. The superior performance of \NAME\ shows that the dynamic pipeline selection has a positive impact agnostic of the underlying KG and dataset.
This also answers our overall research question.

\begin{table}[b]
\caption{Overall performance comparison of static and dynamic pipelines for the KG information extraction task.
}
\centering
\resizebox{.8\columnwidth}{!}{
\begin{tabular}{lccccc}
\toprule
\multirow{2}{*}{\textbf{System}} &
  \multirow{2}{*}{\textbf{Dataset}} &
  \multirow{2}{*}{\textbf{\begin{tabular}[c]{@{}c@{}}Knowledge\\ Graph\end{tabular}}} &
  \multicolumn{3}{c}{\textbf{Performance}} \\ \cline{4-6} 
                              &             &      & \textbf{P} & \textbf{R} & \textbf{F1} \\ \hline
T2KG~\cite{t2kg}                  & WebNLG           & DBpedia         & 0.133      & 0.140      &    0.135          \\ \midrule
\multirow{1}{*}{Frankenstein (Static)~\cite{frankenstein}} & WebNLG           & DBpedia         & 0.177      & 0.189      &    0.181         \\ \midrule
\multirow{2}{*}{\NAME \ (Static)}       & WebNLG           & DBpedia         & 0.210      & 0.225      & 0.215       \\
                              & COV-triples           & ORKG         &  0.403     & 0.423      &  0.413       \\ \midrule\midrule
\multirow{2}{*}{Frankenstein (Dynamic)~\cite{frankenstein}} & WebNLG      & DBpedia  & 0.199      & 0.208      & 0.203       \\
                             & COV-triples & ORKG & 0.403      & 0.424      & 0.413       \\ \midrule
\multirow{2}{*}{\NAME \ (Dynamic)}      & WebNLG      & DBpedia  & \textbf{0.287}      & \textbf{0.307}      & \textbf{0.297}       \\
                              & COV-triples & ORKG & \textbf{0.411}      & \textbf{0.437}      & \textbf{0.424}       \\ \bottomrule
\end{tabular}
}
\label{tab:pipeline-dynamic-performance}
\end{table}

\subsection{Ablation Studies}
\NAME\ and baselines render relatively low performance on all the employed datasets. Hence, in the ablation studies our aim is to provide a holistic picture of underlying errors, collective success, and failures of the integrated components. 

In the first study, we calculate the proportion of errors in \NAME. The modular architecture of the proposed framework allows us to benchmark each component independently. We consider the erroneous cases of \NAME\ on the test set of the WebNLG dataset. We calculate the performance (F1 score) of the \NAME\ dynamic pipeline (\cf Table~\ref{tab:pipeline-dynamic-performance}) at each step in the pipeline.
The results show that the coreference resolution components caused 21.54\% of the errors, 33.71\% are caused by text triple extractors, 18.17\% by the entity linking components, and 26.58\% are caused by the relation linking components.

We conclude that the text triple extractor components contribute to the largest chunk of the errors over DBpedia. 
One possible reason for their limited performance is that open-domain information extracting components were not initially released for the KG information extraction task. Also, these components do not incorporate any schema or prior knowledge to guide the extraction. We observe that the errors mainly occur when the sentence is complex (with more than one entity and predicate), or relations are not explicitly mentioned in the sentence. 
We further analyze the text triple extractor errors. The error analysis at the level of the triple subject, predicate, and object showed that most errors are in predicates (40.17\%) followed by objects (35.98\%) and subjects (23.85\%).

\paragraph{\textbf{Further Analysis}}
Aiming to understand why IE pipelines perform with low accuracy, we conduct a more in-depth analysis per IE task. In the first analysis, we evaluated each component independently on the WebNLG dataset. Researchers~\cite{derczynski2015analysis,singh2019qaldgen} proposed several criterion for micro-benchmarking tools/components for KG tasks (entity linking, relation linking, etc.) based on the linguistic features of a sentence. We motivate our analysis based on the following:

\RNum{1}) {\it Text Triple Extraction}: We consider the number of words (wc) in the input sentence (a sentence is termed by \qq{simple} with average word length of 7.41~\cite{frankenstein}. Sentences with higher number of words than seven are complex sentences). Furthermore, having a comma in a sentence (sub-clause) to separate clauses is another factor. Atomic sentences (\eg \emph{"cats have tails"}) are a type of sentence that also affects triples extractors' behavior. Moreover, nominal relation as in \emph{"Durin, son of Thorin"} is another impacting factor on the performance. Uppercase and lowercase mentions of the words (\ie correct capitalization of the first character and not the entire word) in a sentence are standard errors for entity linking components. We consider this as a micro-benchmarking criteria. 

\RNum{2}) {\it Coreference Resolution}: We focus on the length of the coreference chain (\ie the number of aliases for a single mention). Additionally, the number of clusters is another criterion in the analysis. A cluster refers to the groups of mentions that require disambiguation (\eg \emph{"mother bought a new phone, she is so happy about it"} where the first cluster is \emph{mother} \rarrow\ \emph{she} and the second is \emph{phone} \rarrow\ \emph{it}). The presence of proper nouns in the sentence is studied as well as acronyms. Furthermore, the demonstrative nature of the sentence is also observed as a factor. Demonstrative sentences are the ones that contain demonstrative pronouns (this, that, etc.).

\RNum{3}) {\it Entity Linking}: The number of entities in a sentence (e=1,2) is a crucial observation for the entity linking task. Capitalization of the surface form is another criterion for micro-benchmarking entity linking tools.
An entity is termed as an explicit entity when the entity's surface form in a sentence matches the KG label. An entity is implicit when there is a vocabulary mismatch. For example, in the sentence \emph{"The wife of Obama is Michelle Obama."}, the surface form \emph{Obama} is expected to be linked to \texttt{dbr:Barack\_Obama} and considered as an implicit entity~\cite{singh2019qaldgen}.
The last linguistic feature is the number of words (w) in an entity label (\eg \emph{The Storm on the Sea of Galilee} has seven words). 

\RNum{4}) {\it Relation Linking}: Similar to the entity linking criteria, we focus on the number of relations in a sentence (rel=1,2). The type of relation (\ie explicit, or implicit) is another parameter. Covered relation (sentences without a predicate surface form) is also used as a feature for micro-benchmarking: \emph{"Which companies have launched a rocket from Cape Canaveral Air Force station?"} where the \texttt{dbo:manufacturing} relation is not mentioned in the sentence. Covered relations highly depend on common sense knowledge (i.e., reasoning) and the structure of the KG~\cite{singh2019qaldgen}. Lastly, the number of words (w$<$=N) in a predicate surface form is also considered.

\begin{figure}[p]
  \centering
  \subfigure[F1 score heatmap of the EL task]
  {\includegraphics[width=.8\columnwidth]{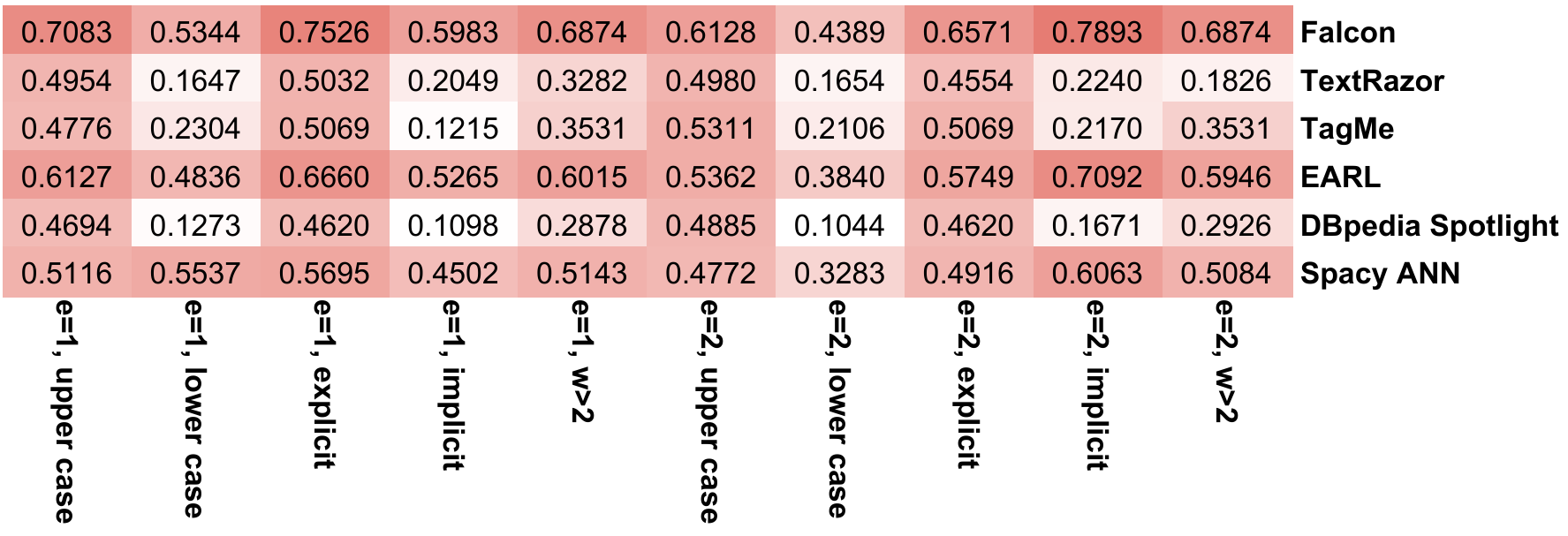}}~
  \newline
  \subfigure[F1 score heatmap of the Text TE task]{\includegraphics[width=.8\columnwidth]{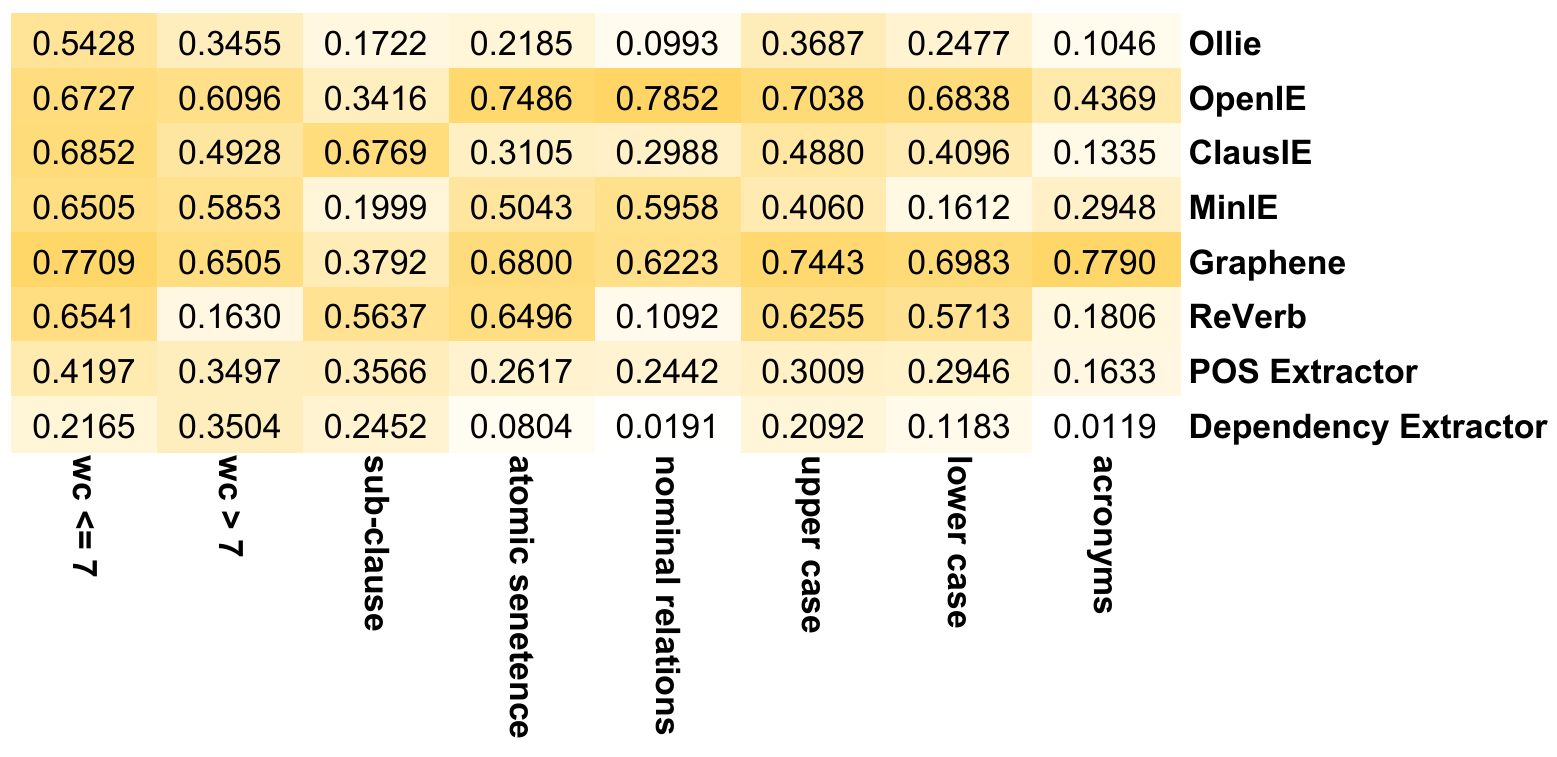}}
  \newline
  \subfigure[F1 score heatmap of the CR task]{\includegraphics[width=.8\columnwidth]{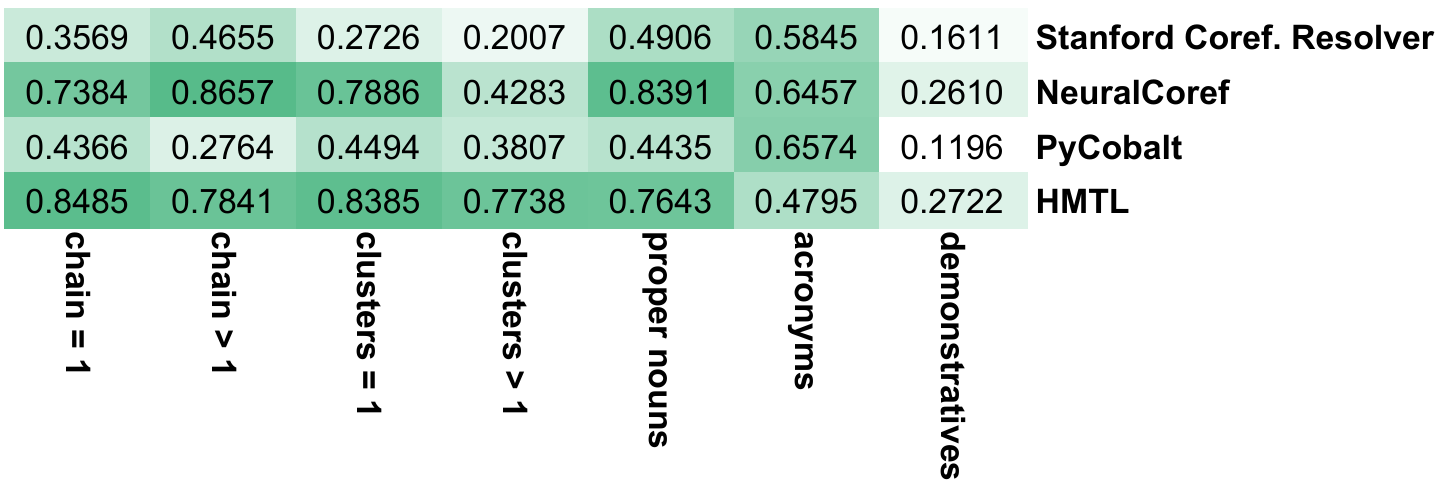}}~
  \newline
  \subfigure[F1 score heatmap of the RL task]{\includegraphics[width=.8\columnwidth]{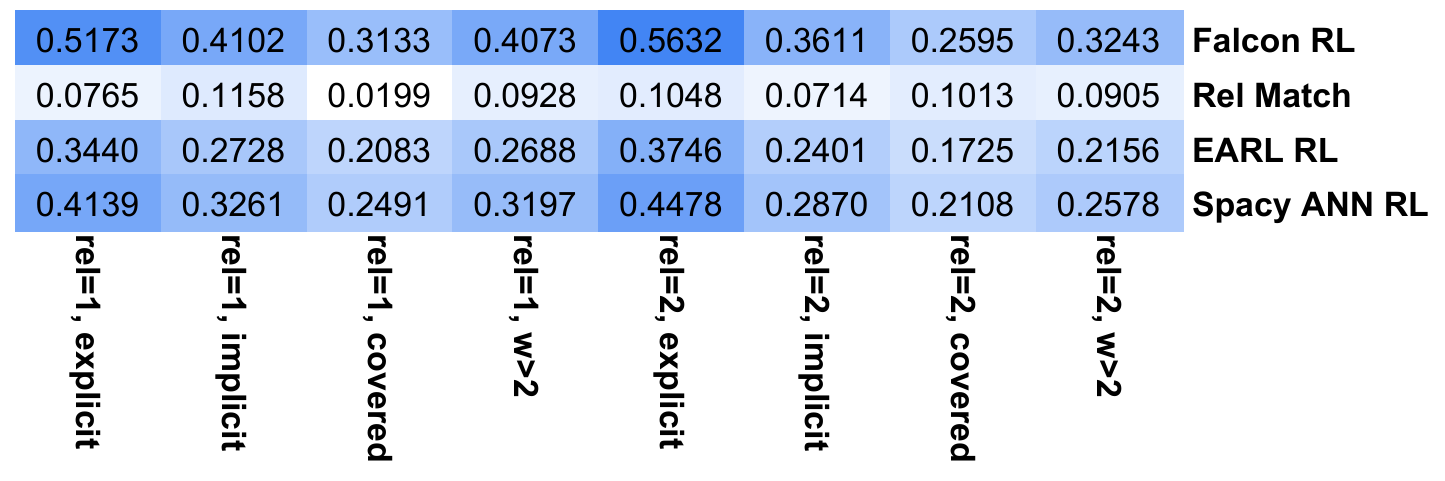}}
  \caption{Comparison of F1 scores per component for different IE tasks based on the various linguistic features of an input sentence (number of entities, word count in a sentence, implicit vs. explicit relation, etc.). Darker colors indicate a higher F1 score.}
  \label{fig:heatmaps}
\end{figure}

Figure~\ref{fig:heatmaps} illustrates micro-benchmarking of various \NAME\ components per task. We observe that across IE tasks, the F1 score of the components varies significantly based on the sentence's linguistic features. In fact, there exist no single component which performs equally well on all the micro-benchmarking criteria. This observation further validates our hypothesis to design \NAME\ for building dynamic information extraction pipelines based on the strengths and weaknesses of the integrated components. We also note in Figure~\ref{fig:heatmaps} that all the CR components report limited performance for the demonstrative sentences (\emph{demonstratives}). When there is more than one coreference cluster in a sentence, all other CR components observe a discernible drop in F1 score. The NeuralCoref~\cite{neuralcoref} component performs best for \emph{proper nouns}, whereas PyCobalt~\cite{pycobalt} performs best for the \emph{acronyms} feature (almost being tied by NeuralCoref). In the TE task, Graphene~\cite{graphene} shows the most stable performance across all categories. However, the performance of all components (except Dependency Parser) drops significantly when the number of words in a sentence exceeds seven (wc$>$7). Case sensitivity also affects the performance and all components observe a noticeable drop in F1 score for lowercase entity mentions in the sentence. Similar behavior is observed for entity linking components where case sensitivity is a significant cause of poor performance. When the sentence has one entity and it is implicit (e=1, implicit); all entity linking components face challenges in correctly linking the entities to the underlying KG. Relation linking components also report lower performance for implicit relations.

\section{Discussion}
\label{sec:discussion}
Even though the dynamic pipelines of \NAME\ outperforms static pipelines, the overall performance of \NAME\ and baselines for the KG information extraction task remains low. Our detailed and exhaustive ablation studies suggest that when individual components are plugged together, their individual performance is a major error source. However, this behavior is expected, considering earlier research works in other domains also observe a similar trend. As in 2015 Gerbil framework~\cite{usbeck2015gerbil} and in 2018 Frankenstein~\cite{frankenstein}.
Within two years, the community has released several components dedicated to solving entity linking and relation linking~\cite{falcon,earl,mihindukulasooriya2020leveraging}, which were two loopholes identified by~\cite{frankenstein} for the QA task. 

We observe that state of the art components for information extraction still have much potential to improve their performance (both in terms of runtime and F1 score). It is essential to highlight that some of the issues observed in our ablation study are very basic and repeatedly pointed out by researchers in the community. For instance, Derczynski et al.~\cite{derczynski2015analysis} in 2015, followed by Singh et al.~\cite{frankenstein} in 2018, showed that case sensitivity is a main challenge for EL tools. Our observation in Figure~\ref{fig:heatmaps} again confirms that case sensitivity of entity surface forms remains an open issue even for newly released components. In contrast, on specific datasets such as CoNLL-AIDA, several EL approaches reported F1 scores higher than 0.90~\cite{yang2019learning}, showing that EL tools are highly customized to particular datasets. In a real-world scenario like ours, the underlying limitations of approaches are uncovered. 

\section{Conclusion and Future Work}
\label{sec:conclusion}
In this paper, we presented the \NAME\ approach and framework for information extraction. \NAME\ effectively selects the best possible pipeline for a given input sentence using the sentential contextual features and a state-of-the-art transformer-based classification model. \NAME\ has a service-oriented architecture which is scalable, extensible, reusable, and agnostic of the underlying KG. The core idea of \NAME\ is to combine the strengths of already existing disjoint research for KG information extraction and build a foundation for a platform to promote reusability for the construction of large-scale and semantically structured KGs. Our empirical results suggest that the performance of the individual components directly impacts the end-to-end information extraction accuracy.

This article does not focus on internal system architecture or employed algorithms in a particular IE component to analyze the failures. The focus of the ablation studies is to holistically study the collective success and failure cases for the various tasks. Our studies provide the research community with insightful results over two knowledge graphs, \componentsCount components, \pipelinesCount pipelines. Our work is a step in the larger research agenda of offering the research community an effective way for synergistically combining and orchestrating various focused IE approaches balancing their strengths and weaknesses taking different application domains into account.
We plan to extend our work in the following directions: \rNum{1}) extending \NAME\ to other KGs such as UMLS~\cite{umls} and Wikidata~\cite{wikidata}. \rNum{2}) addressing multilinguality with \NAME, and \rNum{3}) creating high performing RL components. 
\bibliographystyle{splncs04}
\bibliography{references}

\end{document}